\documentclass[sigconf]{acmart}
\usepackage{float}
\usepackage{comment}
\usepackage{nameref}
\setcopyright{none}  
\renewcommand\footnotetextcopyrightpermission[1]{}  %% adds a footer with the workshop title to each page
\acmConference[Augmented Educators and AI(CHI 2025 Workshop)]{Augmented Educators and AI: Shaping the Future of Human-AI Collaboration in Learning on CHI 2025 Workshop}{April 26,2025}{Yokohama, JAPAN}
\begin{document}
\title{On the development of an AI performance and behavioural measures for teaching and classroom management }

\author{Andreea I.Niculescu}
\affiliation{%
  \institution{A*STAR Inst. for Infocomm Research}
  \city{Singapore}
  \country{}}
%\email{andreea-n@i2r.a-star.edu.sg}
\email {andreeaniculescu@sigchi.org}

\author{Jochen Ehnes}
\affiliation{%
  \institution{A*STAR Inst. for Infocomm Research}
  \city{Singapore}
  \country{}}
\email{jehnes@gmail.com}

\author{Chen Yi}
\affiliation{%
  \institution{A*STAR Inst. High Perf. Computing}
  \city{Singapore}
  \country{}}
\email{cheny@ihpc.a-star.edu.sg}

\author{Du Jiawei}
\affiliation{%
  \institution{A*STAR Inst. High Perf. Computing}
  \city{Singapore}
  \country{}}
\email{dujw@cfar.a-star.edu.sg}

\author{Tay Chiat Pin}
\affiliation{%
  \institution{A*STAR Inst. High Perf. Computing}
  \city{Singapore}
  \country{}}
\email{Tay_Chiat_Pin@ihpc.a-star.edu.sg}

\author{Joey Tianyi Zhou}
\affiliation{%
  \institution{A*STAR Inst. High Perf. Computing}
  \city{Singapore}
  \country{}}
\email{joey_zhou@cfar.a-star.edu.sg}

\author{Vigneshwaran Subbaraju}
\affiliation{%
  \institution{A*STAR Inst. High Perf. Computing}
  \city{Singapore}
  \country{}}
\email{Vigneshwaran_Subbaraju@ihpc.a-star.edu.sg}

\author{Teh Kah Kuan}
\affiliation{%
  \institution{A*STAR Inst. for Infocomm Research}
  \city{Singapore}
  \country{}}
\email{teh_kah_kuan@i2r.a-star.edu.sg}

\author{Tran Huy Dat}
\affiliation{%
  \institution{A*STAR Inst. for Infocomm Research}
  \city{Singapore}
  \country{}}
\email{hdtran@i2r.a-star.edu.sg}

\author{John Komar}
\affiliation{%
  \institution{National Institute of Education NTU}
  \city{Singapore}
  \country{}}
\email{john.komar@nie.edu.sg}

\author{Gi Soong Chee}
\affiliation{%
 \institution{National Institute of Education NTU}
 \city{Singapore}
  \country{}}
\email{gi.soongchee@computing.sg}

%\affiliation{%
%  \institution{National Institute of Education NTU}
%  \city{Singapore}
%  \country{}}
%\email{susan.gwee@nie.edu.sg}

\author{Kenneth Kwok}
\affiliation{%
  \institution{A*STAR Inst. High Perf. Computing}
  \city{Singapore}
  \country{}}
\email{kenkwok@ihpc.a-star.edu.sg}

%%adds a footer with the author infor to each page. 
%%adds a footer with the author infor to each page. 
\renewcommand{\shortauthors}{A.I. Niculescu et al.}
\begin{abstract}
This paper presents a two-year research project focused on developing AI-driven measures to analyze classroom dynamics, with particular emphasis on teacher actions captured through multimodal sensor data. We applied real-time data from classroom sensors and AI techniques to extract meaningful insights and support teacher development. Key outcomes include a curated \textbf{audio-visual dataset}, \textbf{novel behavioral measures}, and a proof-of-concept \textbf{teaching review dashboard}. An initial evaluation with eight researchers from the National Institute for Education (NIE) highlighted the system's clarity, usability, and its non-judgmental, automated analysis approach--which reduces manual workloads and encourages constructive reflection. Although the current version does not assign performance ratings, it provides an objective snapshot of in-class interactions, helping teachers recognize and improve their instructional strategies. Designed and tested in an Asian educational context, this work also contributes a culturally grounded methodology to the growing field of AI-based educational analytics.

\end{abstract}

\maketitle
\flushbottom
{\small
\noindent ©  This paper was adapted for the \textit{CHI 2025 Workshop on Augmented Educators and AI: Shaping the Future of Human and AI Cooperation in Learning},
held in Yokohama, Japan on April 26, 2025. This work is licensed under the Creative Commons Attribution 4.0 International License (CC BY 4.0).
}
\pagestyle{plain} 
\section{Introduction and related work}
Effective teaching is crucial for student learning outcomes, but evaluating teaching performance remains a challenge. Traditional observation methods are often subjective and resource-intensive. AI-driven approaches present a promising solution by leveraging sensor-based observations to objectively quantify teaching behaviors. 

Recent studies have explored the integration of AI and sensor technologies in educational settings. For instance, Holstein et al. \cite{Holstein} developed a mixed-reality teacher awareness tool that enhances student learning by providing real-time analytics to educators. Similarly, Gao et al. \cite{Gao} employed fused sensor data and explainable machine learning models to detect teacher expertise within immersive virtual reality classrooms. These innovations underscore the potential of AI to transform teaching evaluation by offering objective and scalable solutions.

In addition to assessing teaching practices, AI has been applied to monitor student behaviors. Yang and Wang \cite{Fan} introduced the SCB-Dataset3, a benchmark designed for detecting student classroom behaviors, facilitating the development of models that can analyze and interpret student actions in real time.

Dragon et al. \cite{Dragon} investigated how real-time multi-modal sensors, including posture tracking, movement analysis, and facial expression recognition, can provide insights into student affect and learning outcomes. Their study highlighted the potential of combining human observation with automated sensing to enhance educational research and adaptive learning technologies.

To monitor classroom behavior, detecting audio and visual cues is essential. For audio, deep and recurrent neural network (RNN) architectures \cite{Cosbey2019_13} have been applied to detect segments of audio recordings corresponding to single voice (e.g., lecture), multiple voices (e.g., pair discussion), and no voice (e.g., thinking) activities \cite{Owens2017_12}, as well as different instructional formats (e.g., lecture, question and answer, group work) \cite{Donnelly2016_14}. A recent study by Kang et al. \cite{Li2019_15} fused speech and language data and used a multi-modal attention-based neural network to classify classroom activities.

In the visual modality, recent work has favored approaches based on the single-stage \textit{“you only look once”} (YOLO) algorithm for recognizing and analyzing classroom behaviors \cite{Shen2023_8}, \cite{Wang2022_17}, \cite{Du2023_18}, due to its good speed-accuracy balance, over the two-stage Faster R-CNN implementations \cite{Ren2017_19}, which, while more accurate, require longer processing times. Examples of classroom behaviors recognized include positive (e.g., focused, reading, writing, raising hand, and standing up) and negative (e.g., sleeping, looking around, and playing with a phone) behaviors. Additional modalities, such as thermal and photoplethysmography analyses, can also be used to recover physiological measures like skin temperature and heart rate from videos, which can inform the analysis of cognitive workload and engagement.

Despite significant advancements in AI-driven teaching evaluations, research that considers cultural contexts-particularly in Asian settings—remains limited. This study addresses this gap through work conducted in Singaporean classrooms, developing AI-based measures tailored to local teaching behaviors, along with an annotated dataset and a  dashboard for self-review. The project offers culturally relevant insights and practical tools, contributing to the broader discourse on AI in education.

\section{Methods}
Our project, carried out at the National Institute of Education (NIE), Nanyang Technological University (NTU) in Singapore, was conducted over a period of two years—from April 2022 to April 2024. Before commencing data collection, we obtained ethics approval from the relevant institutional review boards. This approval was granted in October 2022, six months after the project began, following a rigorous review process to ensure adherence to ethical research standards, including participant confidentiality and data protection.
\subsection{Data collection}
The data collection process took place over three academic semesters and yielded a total of 53 recordings, comprising approximately 90 hours of footage. Individual session lengths ranged from around 30 minutes to just under 3 hours, with an average duration of 1 hour and 38 minutes.

At the beginning of the study, five pilot recordings were conducted to test the recording protocols. These sessions focused on evaluating key setup aspects, particularly the positioning of recording devices—especially the iPhones used to capture participants' facial expressions. Recording students’ facial expressions is essential for understanding their reactions to teacher actions and is closely linked to evaluating teaching quality. 

Insights gained during the pilot phase led to refinements in camera placement, including the addition of a second student-facing webcam to improve visual coverage in the main study. As a result, the pilot sessions differ slightly from the main dataset, as they were conducted prior to the implementation of the final, optimized camera setup. 

In addition, technical issues—primarily involving the ceiling-mounted webcam—resulted in missing or corrupted footage for eight sessions. To maintain data quality, these recordings were excluded, bringing the final dataset to 45 recordings containing approx. 69 hours of video data. 
\subsection{Subject recruitment}
The data were collected during regular classroom lessons at NIE, with teacher volunteers recruited from the faculty and students from their existing classes. Informed consent was obtained from all participants, who were fully briefed on the study’s objectives, data collection procedures, and the intended use of the recordings to ensure ethical compliance.

The final collection comprises video and audio recordings from 12 teachers (11 male, 1 female). Each teacher was invited to contribute up to 5 lessons, though actual contributions varied due to scheduling constraints, absences, and classroom availability. To ensure consistency in the recording environment, all sessions were conducted in a designated experimental classroom—without altering the instructional content or teaching format. This setup enabled the capture of a wide range of authentic teaching scenarios under standardized conditions, yielding a rich resource for analyzing classroom interactions and instructional practices.
\subsection{Experimental classroom setup}
\begin{figure}[h!] % H from the float package forces exact placement
    \centering
    \includegraphics[width=0.9\columnwidth]{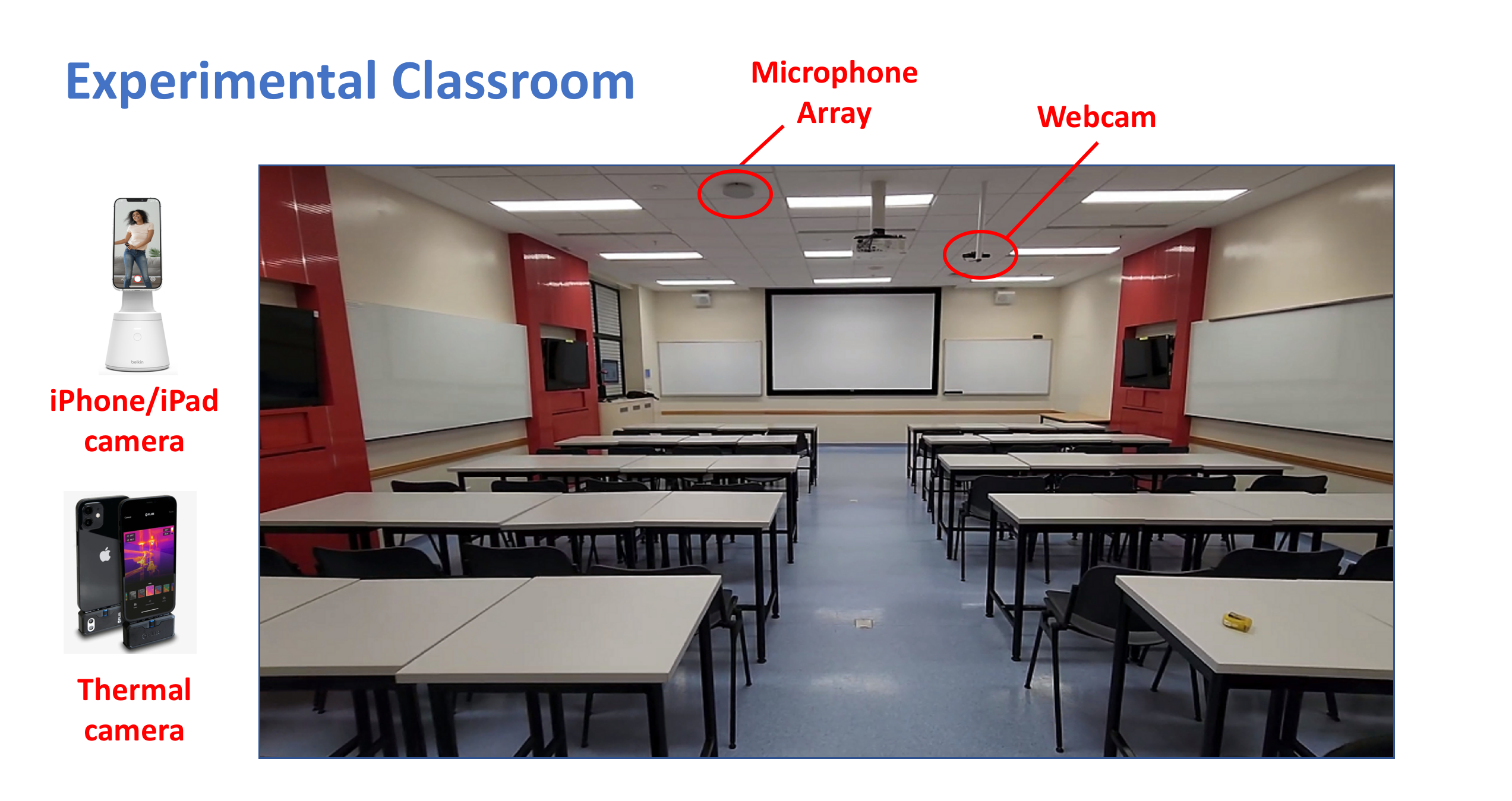} % Adjust width as needed
    \caption{Camera placement in the classroom}
    \label{experimental}
\end{figure}

As shown in Figure~\ref{experimental}, the designated experimental classroom was equipped with a comprehensive array of recording devices designed to capture both teacher and student behaviors. These included a ceiling-mounted microphone array and webcam, a rear-mounted Sony digital camera, an iPad for recording the teacher’s facial expressions, two student-facing webcams, and four iPhones paired with thermal cameras to monitor student activity.

The iPhones were placed on student tables to collect thermal and behavioral data from consistent angles. A synchronization system ensured all devices operated simultaneously, minimizing data inconsistencies.

Capturing high-quality audio was equally critical, particularly since not all microphones were close-talk. Although the classroom was relatively quiet, reverberation posed challenges for distant microphone recordings. To address this, we relied on a robust microphone infrastructure drawing on our previous work in acoustically complex environments. Our experience with setups in noisy and reverberant conditions—such as in-vehicle applications, foyers, and industrial sites—proved valuable in optimizing audio quality for this study \cite{Chong, Robot_Olivia, Robot_Framework, Smart_City, IDA_Interact, TechForFuture}.
\subsection {Software tools for data analysis and labeling}
\subsubsection{Data analysis tools}  To facilitate playback and inspection of the recordings, we developed a specialized tool inspired by multi-track editing software like \textit{Adobe Premiere} and profiling applications such as \textit{Instruments} on macOS.

\begin{figure}[h!] % H from the float package forces exact placement
    \centering
\includegraphics[width=0.9\columnwidth]{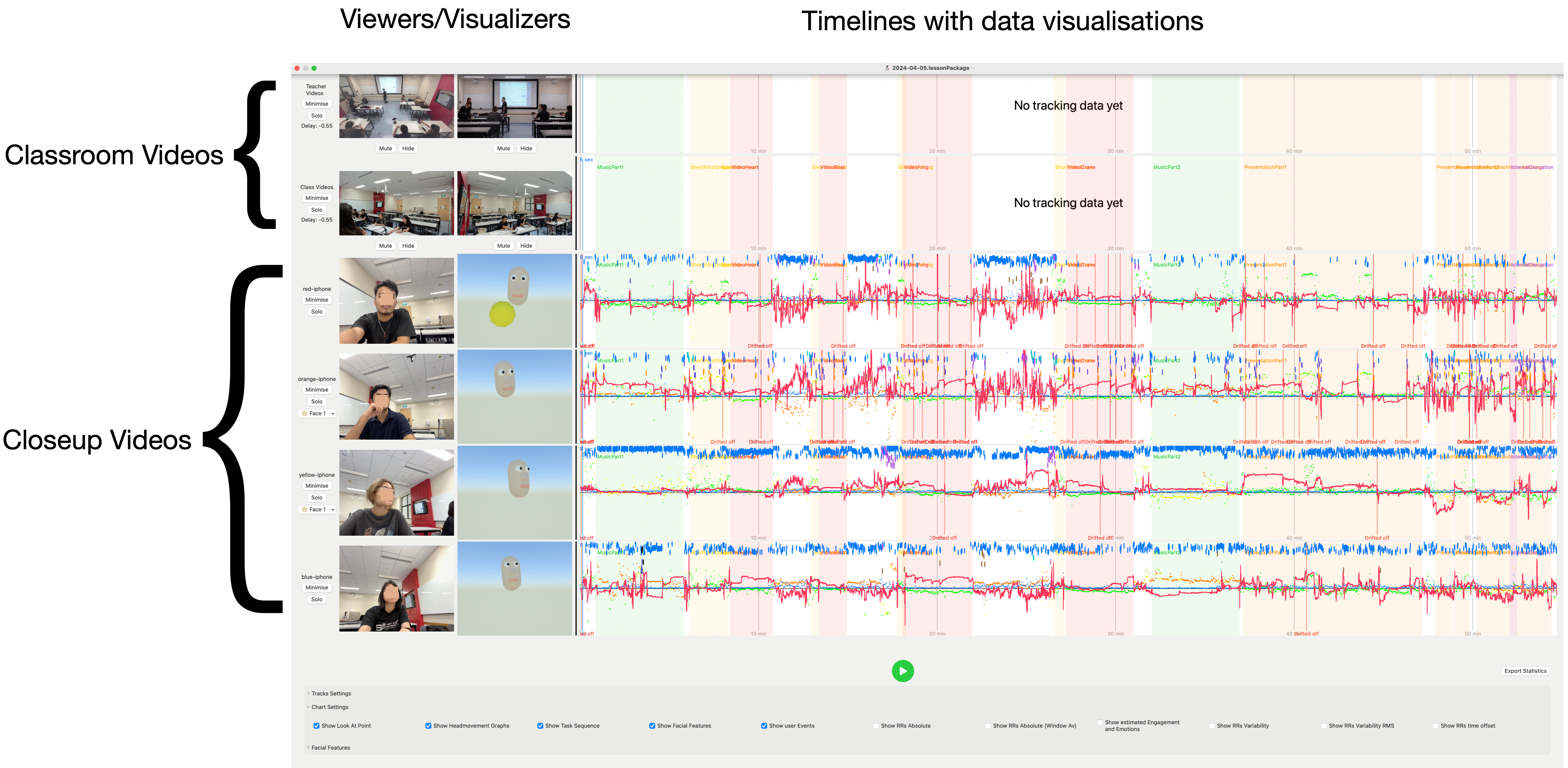} % Adjust width as needed
    \caption{Tracking system}
    \label{tracking}
\end{figure}

The interface consists of multiple rows representing different recording streams. As shown in Figure \ref{tracking}, each row includes a configurable set of video viewers and head-tracking visualizers (on the left), and a timeline view (on the right). The top-left video viewer displays classroom footage recorded from different cameras. Users can hide or maximize individual video viewers for better visibility and mute audio for specific players as needed.

The timeline view allows users to navigate to specific points in the recordings and visualize various behavioral features, such as \textbf{gaze direction}, \textbf{head movements}, \textbf{facial expressions}, and \textbf{user-triggered events}. Zooming options support both detailed, frame-by-frame inspection and a comprehensive overview of the entire session. Playback across all video viewers is synchronized and controlled by a single button—or the space bar—for efficient review.
 
Two cameras positioned at the front recorded the teacher, while two wide-angle cameras focused on the students. To optimize screen space, we combined the two teacher video feeds into a single track and did the same for the student feeds, referred to as "class videos" (both "teacher videos" and "class videos" are grouped under the broader category of "classroom videos"). Teacher tracking data, processed during post-analysis, is visualized within the "teacher videos" track timeline (the topmost stream in the "classroom videos" sequence). Figure \ref{tracking} also shows the default indication that appears when no tracking data has been imported.

Beyond the classroom-wide recordings, we also captured four student videos, providing close-up views of selected students' faces using iPhones. These devices not only recorded video but also tracked and logged head position, gaze direction, and facial features (via ARKit blendshapes\cite{ARkit}). For some recordings, the iPhones additionally captured thermal images and heartbeat data using a chest strap connected to the phones.

Each student video track can display a visualizer showing head-tracking data alongside gaze points (represented by a yellow  pill-shaped ball). Optionally, gaze direction can also be represented using a heatmap visualization. 

The timeline views further depict changes over time enabling users to quickly jump to key events and analyze what was happening in the classroom at those moments. For example, eye-tracking data can indicate whether a student is paying attention to the front of the classroom or is instead focused on their notes. At the same time, facial feature analysis can detect signs of disengagement, such as frequent blinking or yawning. 

\subsubsection{Data annotation}
\begin{figure}[h!] % H from the float package forces exact placement
    \centering
    \includegraphics[width=0.9\columnwidth]{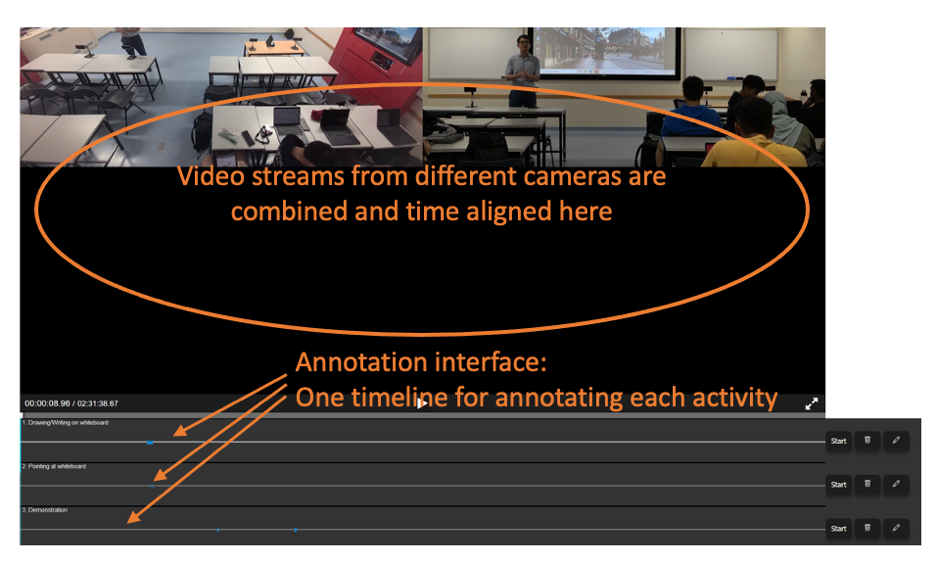} % Adjust width as needed
    \caption{Annotation app}
    \label{annotation}
\end{figure}

To train computer vision models to recognize relevant teaching activities, we first needed to annotate these activities within a training dataset. To support this process, we developed a web-based labeling tool specifically designed to facilitate the annotation of key teaching behaviors (see Figure \ref{annotation}). Due to resource constraints, the labeling focused primarily on low-level teaching actions rather than on complex pedagogical interactions. In total, ten training videos—comprising 13 hours and 42 minutes of footage—were manually annotated for use in training the AI model.

\section{Output results}
\subsection {Audio-Visual annotated data collection}
Building on the data collection and annotation processes described in Sections 2.1 and 2.4.2, the resulting dataset comprises curated, time-synchronized audio-visual recordings of classroom teaching interactions in higher education contexts in Singapore. It integrates footage from multiple capture devices and includes time-aligned teacher actions and student reactions. Teaching activities were annotated using both manual labels and computer vision models, with all annotations saved as plain text files. Student video data is included for context but was not annotated due to resource constraints. The dataset is available to researchers upon request (see \textbf{\nameref{sec:data_collection}}, page 7).
\subsection{AI-Based measures of teaching behavior}

We developed quantifiable measures of teacher behaviors using AI-driven computer vision and speech analysis based on sensor data. Our analysis emphasized non-verbal and paralinguistic features, including teacher location, gestures, use of instructional tools (e.g., slides, whiteboards), and vocal attributes such as loudness, pitch, speaking rate, etc. While content-related descriptors were manually annotated to explore teaching styles, no verbal content was used to train the system. Content-based analysis will be explored in future work, pending Institutional Review Board (IRB) approval.

\subsubsection{Computer vision}
To analyze teacher actions and movement within the classroom, we employed a combination of computer vision models. Teacher position tracking was performed using the YOLOv8 object detection network \cite{jocher2023ultralytics_1} with pretrained weights to detect human locations and poses. The teacher was initially identified based on the highest foot coordinates among all detected humans in the first frame. To maintain consistent tracking, we implemented DeepOCSORT for re-identification (ReID) \cite{maggiolino2023deep_2} and introduced several enhancements to address tracking failures. These included leveraging a history of the teacher’s previous positions, maintaining a student ID list to prevent incorrect reassignments, and implementing an exit detection mechanism to handle cases where the teacher temporarily left the classroom.

As described earlier, teacher actions were labeled using both manual and automatic methods. These actions included \textit{writing on the board}, \textit{pointing at board/screen}, \textit{gesturing at the board}, \textit{hand gestures} (other than those already mentioned), and \textit{slide changes}. To recognize these actions, we applied a pretrained ST-GCN model \cite{yan2018spatial_3}, which extracts skeletal information from video frames. The model achieved classification accuracies of 83.33\% for writing on the board, 65.06\% for pointing at the board/screen, and 88.57\% for gesturing at the board. Slide transitions were detected using the \textit{"Segment Anything"} visual foundation model \cite{kirillov2023segment_4, zhang2023faster_5} with MobileSAM for efficiency. Additionally, hand-waving gestures were recognized using pretrained \textit(mmpose
weights \cite{mmpose2020_6}, which extracted hand coordinates to determine motion patterns.

All detected actions and their corresponding timestamps were stored in a structured JSON format, ensuring a comprehensive and well-organized dataset for further analysis.
\subsubsection{Speech analysis}
Our framework assessed teacher vocal dynamics by extracting \textbf{acoustic features} and deriving higher-level vocal metrics, collectively referred to as \textbf{speaking style}. Using Praat \cite{Praat}, we extracted features such as \textit{loudness}, \textit{pitch}, \textit{rhythm}, \textit{intonation}, and \textit{prosody}, which informed measures like \textit{speaking rate}, \textit{vocal clarity}, and \textit{voice monotony}. In Singapore, clear speech and native English accents—especially British—are preferred over local varieties in public settings \cite{VoiceAccents}.

To quantify speech activity, we implemented a two-level segmentation system that distinguished teacher speech from silence while reducing noise-related errors. The \textbf{coarse level} (60-second intervals) provided an overview of vocal dynamics, while the \textbf{fine level} (10-second intervals) offered detailed analysis. Utterances under 125 milliseconds were excluded as non-informative. We categorized extracted data into three feature groups:

\begin{itemize}
\item \textbf{Statistical summaries}: e.g., \textit{loudness} (dB), \textit{pitch} variability, \textit{formant frequencies}, and \textit{voicing probability}.
\item \textbf{Contextual features}: e.g., \textit{utterance length}, \textit{pause duration}, and \textit{speaking rate}.
\item \textbf{Linguistic features}: e.g., \textit{vocal clarity} (cepstral peak prominence, jitter, shimmer), \textit{intonation}, and \textit{prosody}.
\end{itemize}

To evaluate speaking performance, we applied a Multiple Criteria Decision Making (MCDM) model \cite{StanleyZionts1979}, scoring features on a normalized [0,100] scale using two approaches: \textit{Equal Weights} (EW) and \textit{Reciprocal Standard Deviation Weights} (RW), where more variable features contribute less to the final score.

This dual-layered analysis provides feedback at both overview and detailed levels: coarse-grained trends in \textbf{speaking style} and fine-grained insights into key \textbf{acoustic features}. These insights support teacher reflection on their vocal delivery and overall classroom communication.

\subsection{\textit{Teacher Trainer} dashboard}
The \textit{Teacher Trainer} dashboard prototype is a proof-of-concept visualization of the measures created to support teacher professional development. The dashboard consists of two screens: \textbf{summary} and \textbf{review}.

\subsubsection{Summary screen} The \textbf{summary screen} offers an overview of \textbf{teaching behaviors (1)}, \textbf{spatial movements (2)}, and \textbf{speech \& movement characteristics (3)}. Due to its large size, we will present its components individually rather than displaying the entire interface at once.

\begin{figure}[h!] % H from the float package forces exact placement
    \centering
\includegraphics[width=0.9\columnwidth]{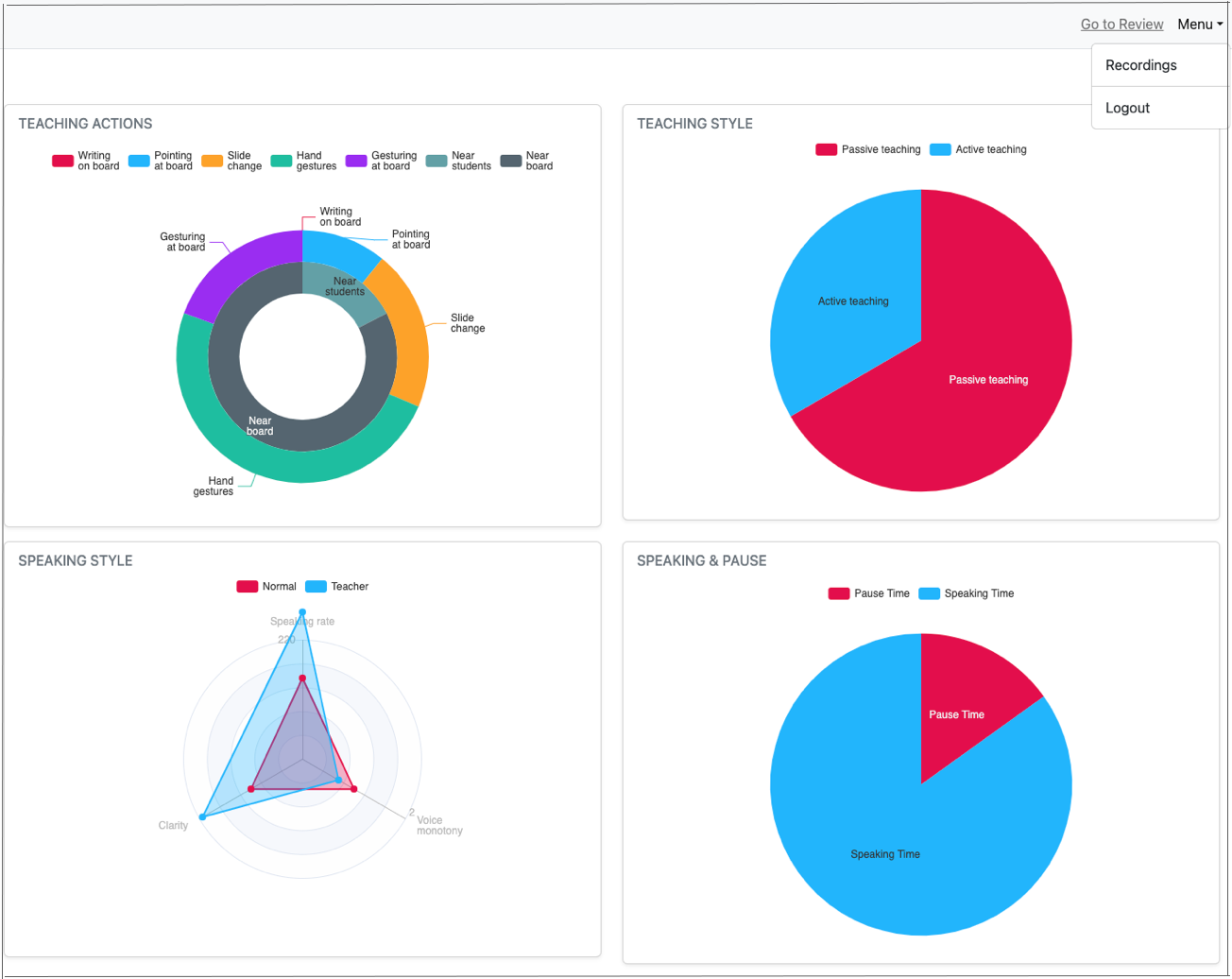} % Adjust width as needed
    \caption{Summary screen (1) - teaching behaviors}
    \label{screen1}
\end{figure}

Figure \ref{screen1} depicts four distinct panels positioned on the right side of the \textbf{summary screen} representing \textbf{teaching behaviors}:
\begin{itemize}
\item \textbf{Teaching Actions (top left)}: displays a donut chart; the outer ring shows the proportion of time the teacher spends performing various actions (e.g., hand gestures, writing on the board, pointing at the board, and gesturing at the board); the inner ring breaks this down further by showing time spent near the board versus near the students.

\item \textbf{Teaching Style (top right)}: represents the balance between active and passive teaching methods, inferred from the tea- cher’s observed actions. Passive teaching generally encourages convergent thinking—guiding students toward a single correct answer—while active teaching promotes lateral thinking, helping students connect concepts to real-world contexts \cite{rotgans2011role}. As mentioned earlier, these data were obtained exclusively through manual annotations.

\item \textbf{Speaking Style (bottom left)}: captures vocal characteristics benchmarked against literature-based norms. The norm for \textbf{speaking rate} is set at 140 words per minute (wpm), within the effective lecture delivery range of 125–160 wpm \cite{Tauroza}. For \textbf{vocal clarity}, we adopt a Speech Transmission Index (STI) threshold of 0.75; scores above 0.5 are acceptable, and those above 0.75 are optimal \cite{Steeneken}, \cite{Lecesse}. For \textbf{voice monotony}, we set a norm of 1.0 based on intonation scores: values below 0.4 indicate monotonous speech, around 1.0 reflect average expressiveness, and values above 1.6 suggest a lively delivery \cite{Büllow}.

\item \textbf{Speaking \& Pauses (bottom right)}: visualizes the previously discussed speaking-to-pausing ratio, offering a quick interpretation of conversational dominance.
\end{itemize}
\begin{figure}[h!] % H from the float package forces exact placement
    \centering    \includegraphics[width=0.9\columnwidth]{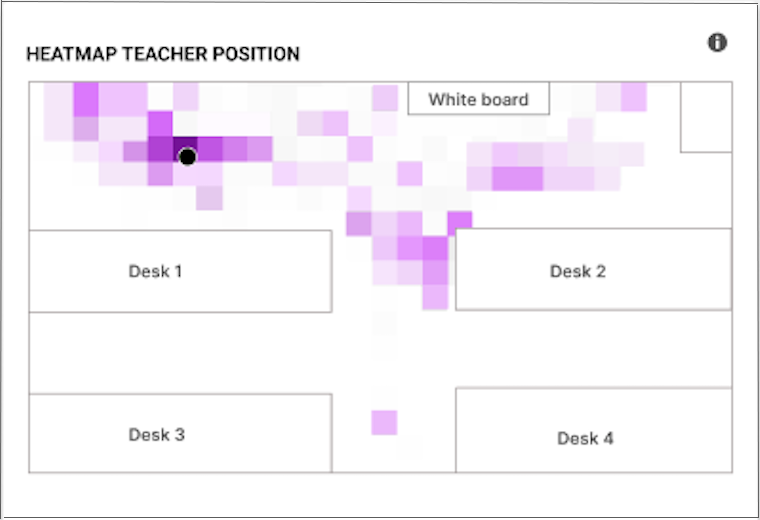} % Adjust width as needed
    \caption{Summary screen (2) - spatial movement}
    \label{screen2}
\end{figure}
 Positioned in the upper left section of the \textbf{summary screen}, Figure \ref{screen2} visualizes the teacher's 
 \textbf{spatial movement} throughout the class. It employs a heatmap to represent the areas where the teacher spends most of the time, with warmer colors indicating higher activity levels. This visualization provides insights into classroom dynamics, such as whether the teacher remains primarily near the board, moves frequently among students, or exhibits a balanced distribution of movement. Analyzing these patterns allows educators to reflect on their movement and consider its potential impact on student engagement and interaction.

\begin{figure}[h!] % H from the float package forces exact placement
    \centering
    \includegraphics[width=0.9\columnwidth]{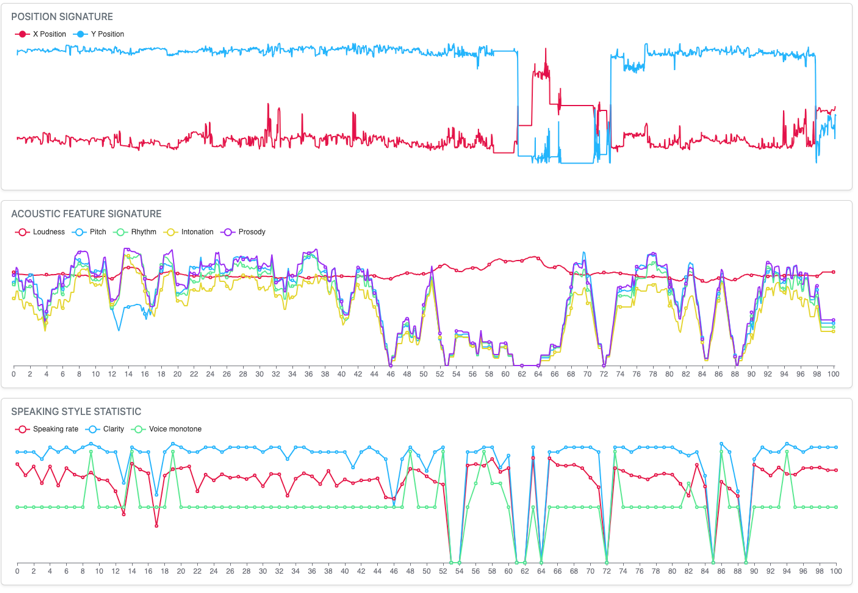} % Adjust width as needed
    \caption{Summary screen (3) - speech \& movement characteristics}
    \label{screen3}
\end{figure}

 Figure \ref{screen3}, located in the lower left section of the \textbf{summary screen}, presents the final panel: a dynamic visualization of the \textbf{teacher’s X-Y position} throughout the session. The X-axis represents movement along the left-right direction, while the Y-axis reflects movement along the front-back axis of the classroom. This visualization tracks how the teacher navigates the space during the lesson, offering insights into movement patterns and engagement strategies. It builds on the positional data discussed earlier, now rendered as time-series movement paths.
 
 Additionally, this panel displays key speech characteristics, such as \textbf{low-level acoustic features} and \textbf{higher-level speaking style characteristics}. These visualizations rely on metrics described in Section 3.2.2. 
\subsubsection{Review screen}
The \textbf{review screen} offers an interactive platform for exploring teaching sessions through synchronized \textbf{video playback} and \textbf{behavioral features}. This integration enables educators to revisit their instructional delivery while simultaneously examining position, gesture, and vocal features.

\begin{figure}[h!] % H from the float package forces exact placement
    \centering
    \includegraphics[width=0.9\columnwidth]{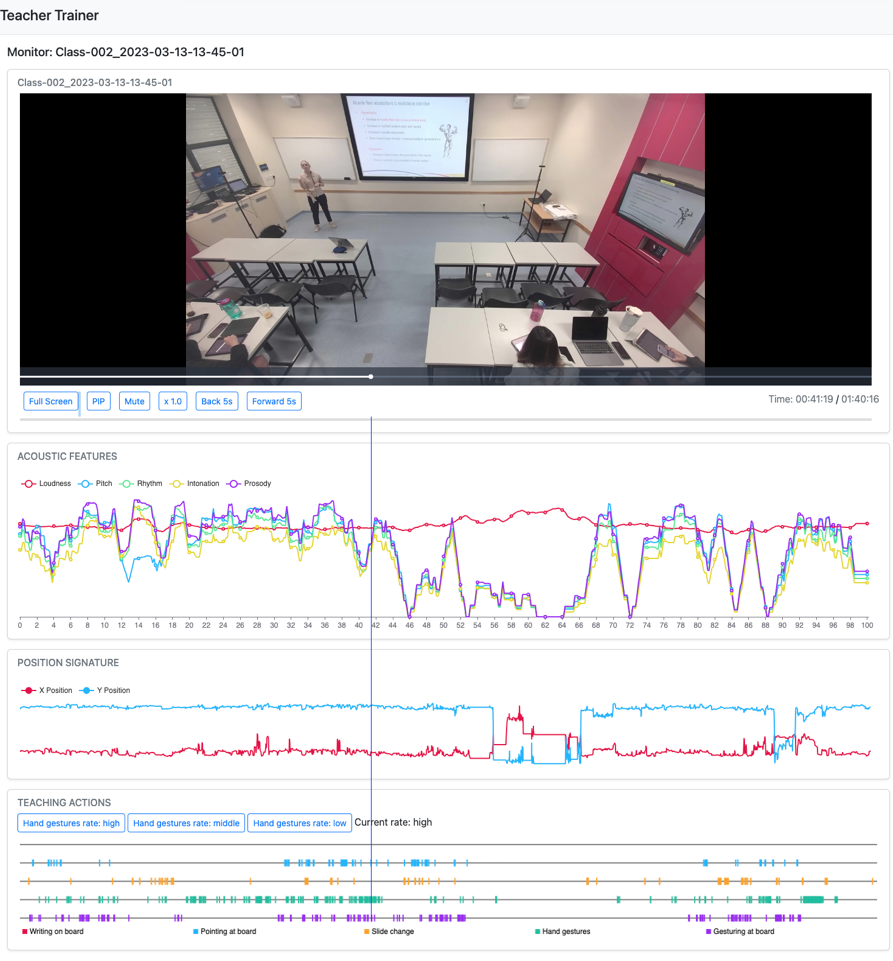} % Adjust width as needed
    \caption{Review screen (left) - video playback \& position and verbal signatures }
    \label{screen4}
\end{figure}

The review screen interface is divided into two parts: a left side and a right side. Figure~\ref{screen4} depicts the left side of the review screen, which contains the \textbf{video playback} synchronized with visualizations of the teacher’s \textbf{position}, \textbf{acoustic features}, and \textbf{speaking style}. These panels mirror core information from the \textbf{summary screen}, but here they are time-synchronized with the video and continuously updated via a dynamic vertical blue bar. This setup allows users to directly correlate observed behaviors with the associated vocal and movement patterns.

Playback controls include flexible navigation features: users can adjust speed (1x–3x), fast-forward or rewind, and switch between full-screen and Picture-in-Picture (PIP) mode. The PIP function supports uninterrupted review by keeping the video accessible while navigating other panels.

\begin{figure}[h!] % H from the float package forces exact placement
    \centering
    \includegraphics[width=0.9\columnwidth]{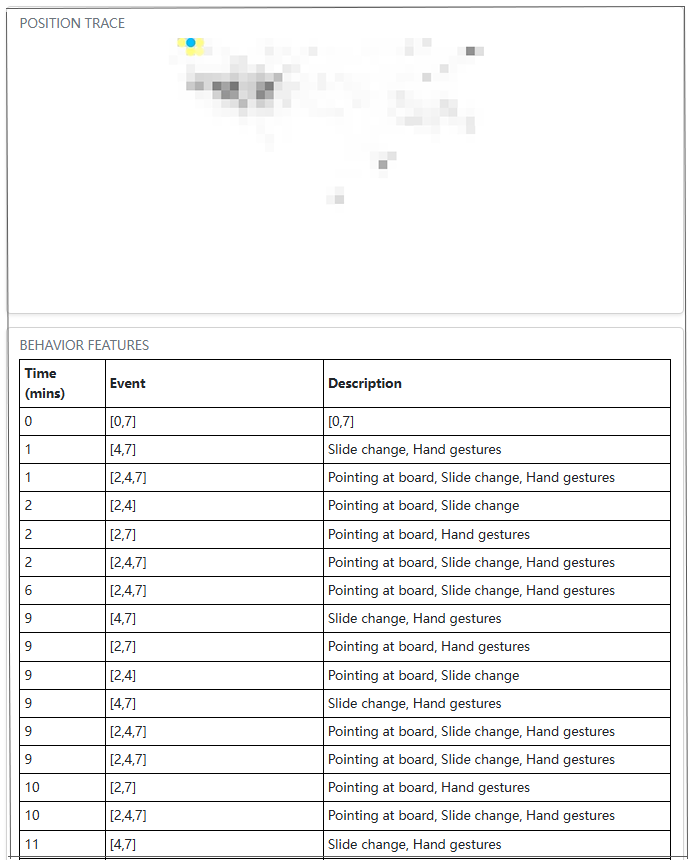} % Adjust width as needed
    \caption{Review screen (right) - latest teacher movement and behavioral features}
    \label{screen5}
\end{figure}
 
As shown in Figure \ref{screen5}, the right side of the review screen complements the summary by visualizing the teacher’s position trace over the past minute, offering a time-sensitive snapshot of spatial dynamics that highlights areas of concentrated movement or stationary behavior.

Below the position trace, a time-stamped textual summary of teaching events is shown. These \textbf{behavioral features} include key actions such as board writing, gestures, pauses, and changes in speaking style or positioning. These annotations help teachers quickly locate pivotal moments and explore opportunities for instructional improvement.

A video demonstration of the teacher trainer dashboard can be accessed at: https://vimeo.com/1076482827 . 

\subsubsection{Evaluation} 
The prototype dashboard underwent an informal evaluation by eight researchers from the National Institute of Education (NIE), affiliated with the Learning Strategy Group (Centre for Innovation in Learning) and the Learning Sciences and Innovation Research Programme (Centre for Research in Pedagogy and Practice, Office of Education Research).

Overall, the feedback was very positive. Evaluators praised the dashboard’s non-judgmental approach to self-review and its potential as a teacher-training aid. They also highlighted the advantages of automated audio–video analysis and event detection—contrasting it with their current manual workflows—and expressed strong interest in further testing. While these early insights are encouraging, a more standardized evaluation protocol is needed to systematically assess the dashboard’s usability and effectiveness.

\begin{figure}[h!] % H from the float package forces exact placement
    \centering
    \includegraphics[width=0.9\columnwidth]{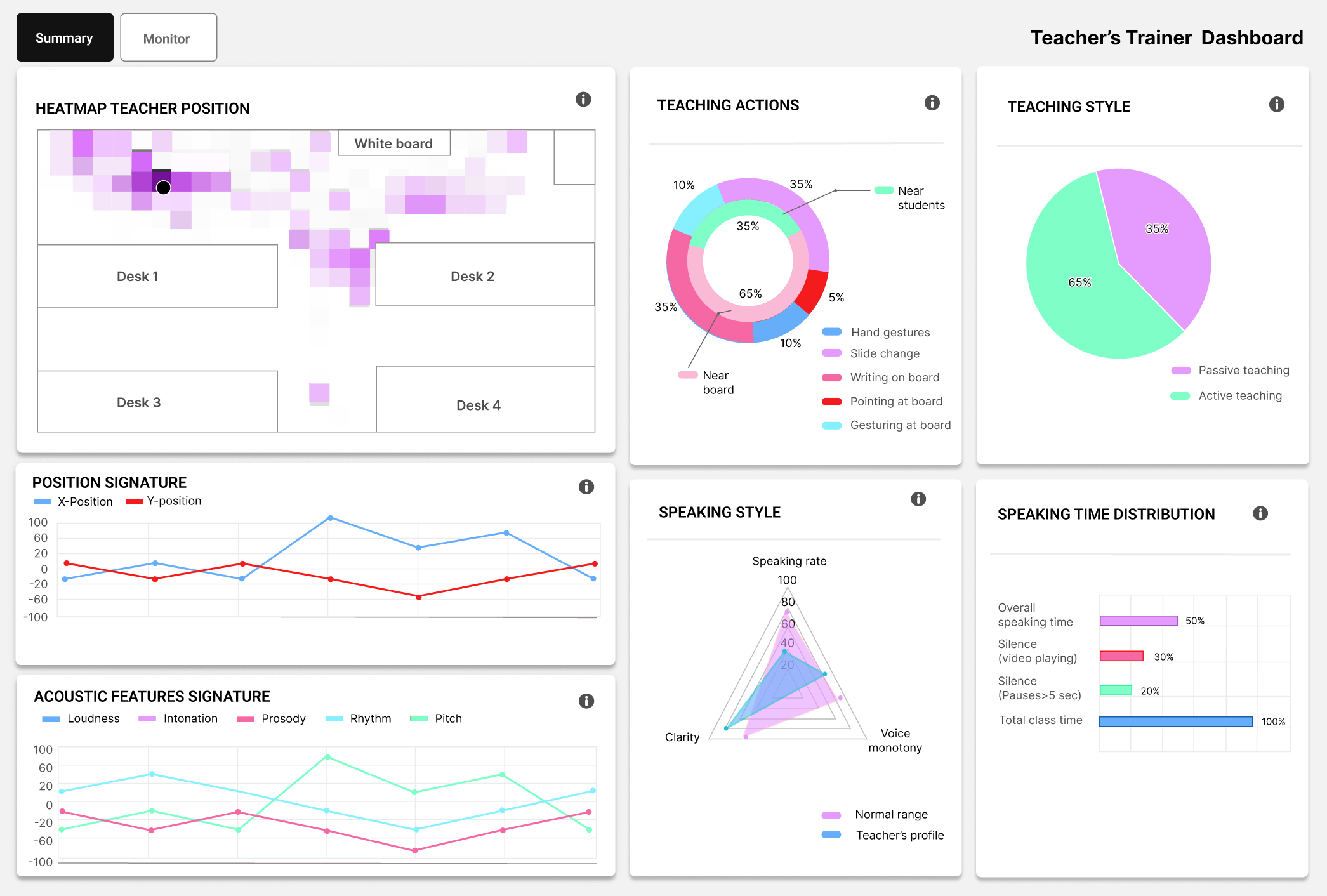} % Adjust width as needed
    \caption{Summary screen}
    \label{summary}
\end{figure}

\begin{figure}[h!] % H from the float package forces exact placement
    \centering
    \includegraphics[width=0.9\columnwidth]{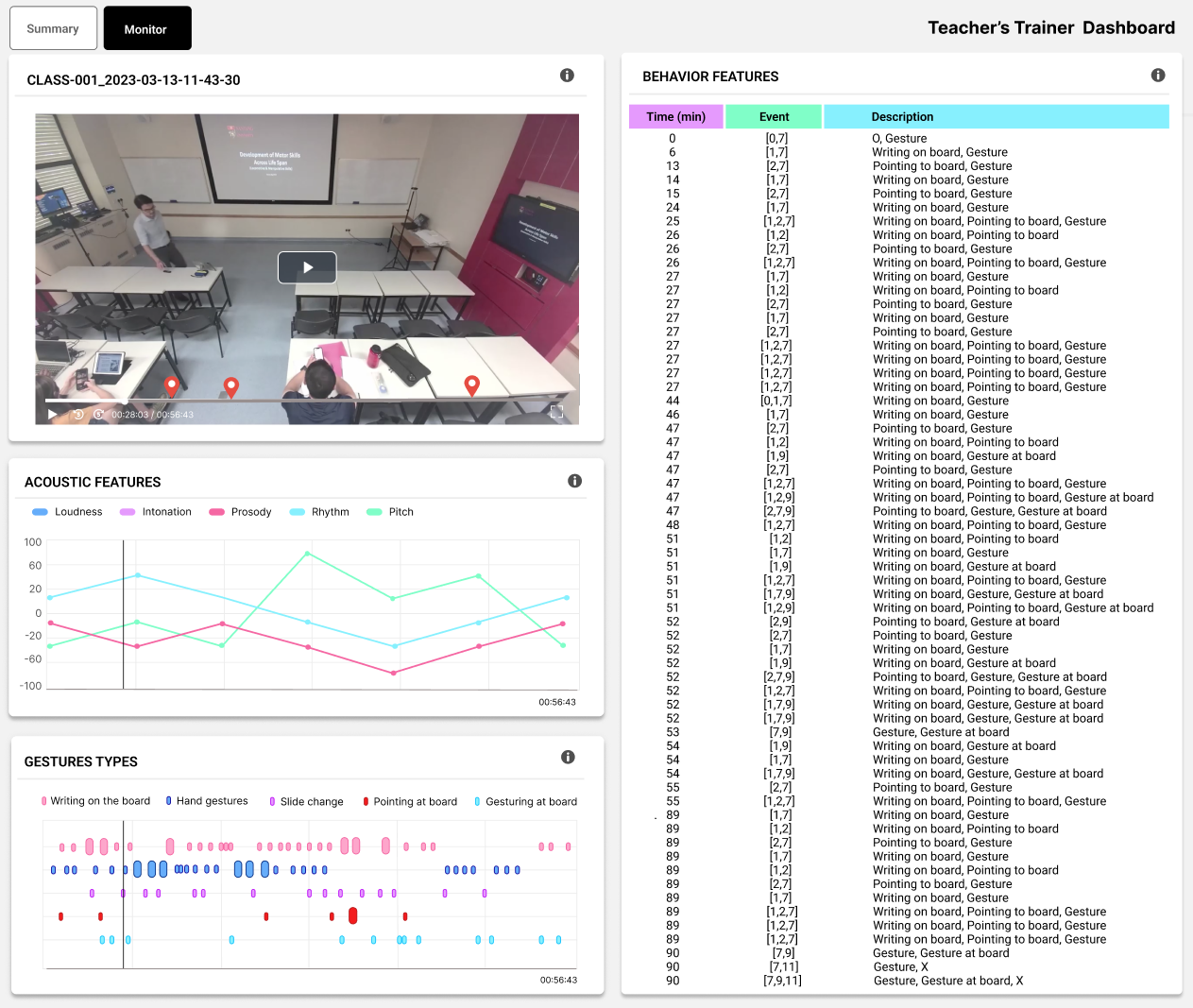} % Adjust width as needed
    \caption{Monitor Screen}
    \label{monitor}
\end{figure}

Based on this feedback, our team refined the screen designs to enhance clarity and user experience (see Figures \ref{summary} and \ref{monitor}).

%Although the current prototype does not rate performance, it provides a clear, objective snapshot of classroom behavior—empowering teachers to reflect on, learn from, and continuously improve their own practice.

\section {Conclusion and future work}
This study explored using AI to generate quantifiable insights into teacher behavior via multimodal sensor data. With our curated dataset, we identified key visual and vocal indicators of classroom practice and showed how AI can provide objective, real-time feedback on non-verbal instructional strategies.

The prototype dashboard—built using computer vision and speech processing techniques—was positively received by teacher evaluators for its intuitive and non-judgmental design. While it does not generate performance ratings, it provides a structured visual summary of teaching behavior, supporting meaningful reflection and ongoing professional development.

The system’s ability to automatically detect actions such as writing on the board, gesturing, and classroom movement enabled teachers to revisit and assess critical moments in their lessons. However, evaluators highlighted the importance of adding higher-order markers—such as questioning, prompting, and encouraging participation—to better capture instructional intent.

Although this phase focused primarily on teacher behavior, the framework is readily extensible to include student-related metrics, such as engagement and cognitive load. This would support a more holistic understanding of classroom dynamics. Future work will focus on expanding behavioral coverage, enabling real-time feedback, and integrating content-level analysis, subject to ethical approval.

Despite current limitations, this project demonstrates the potential of AI-driven classroom behavior analysis and offers valuable contributions to educational analytics: an annotated multimodal dataset, novel behavioral measures, and a working prototype of behavior tracking. These contributions are particularly relevant in Asian educational contexts, where pedagogical norms and evaluation criteria may differ from Western frameworks.

\begin{acks} 
This research was supported by NIE and A*STAR. We thank the participating teachers and students for their contributions.
\end{acks}
\section*{Data collection information}
\phantomsection
\label{sec:data_collection}
For those interested in our data collection, please contact Dr. John Komar at  
\texttt{john.komar@nie.edu.sg}

\bibliographystyle{ACM-Reference-Format}
\bibliography{nie}
\end{document}